\documentclass[journal]{IEEEtran}
\usepackage{amsmath}
\usepackage{mathrsfs}
\usepackage{amsfonts}
\usepackage[numbers,sort&compress]{natbib}
\usepackage{amssymb}
\usepackage{bm}
\usepackage{enumerate}
\usepackage[pdftex]{graphicx}
\usepackage{lineno,hyperref}
\emergencystretch=\maxdimen
\ifCLASSINFOpdf
\else
\fi

\begin{document}
%
\title{Simultaneously Color-Depth Super-Resolution with Conditional Generative Adversarial Network}
%
%
%

\author{Lijun~Zhao,
        Jie~Liang,~\IEEEmembership{Senior~Member,~IEEE,}
        Huihui~Bai,~\IEEEmembership{Member,~IEEE,}
        Anhong~Wang,~\IEEEmembership{Member,~IEEE,}
        and~Yao~Zhao,~\IEEEmembership{Senior~Member,~IEEE}
\thanks{L.  Zhao, H. Bai, Y. Zhao are with Institute Information Science, Beijing Jiaotong University, Beijing, 100044, P. R. China, e-mail: {15112084, hhbai, yzhao}@bjtu.edu.cn.}
\thanks{J. Liang is with School of Engineering Science, Simon Fraser University, ASB 9843, 8888 University Drive, Burnaby, BC, V5A 1S6, Canada, e-mail:jliang@sfu.ca}
\thanks{A. Wang is with Institute of Digital Media \& Communication, Taiyuan University of Science and Technology, Taiyuan, 030024, P. R. China, e-mail:wah\_ty@163.com}}

\maketitle

\begin{abstract}
Recently, Generative Adversarial Network (GAN) has been found wide applications in style transfer, image-to-image translation and image super-resolution. In this paper, a color-depth conditional GAN is proposed to concurrently resolve the problems of depth super-resolution and color super-resolution in 3D videos. Firstly, given the low-resolution depth image and low-resolution color image, a generative network is proposed to leverage mutual information of color image and depth image to enhance each other in consideration of the geometry structural dependency of color-depth image in the same scene. Secondly, three loss functions, including data loss, total variation loss, and 8-connected gradient difference loss are introduced to train this generative network in order to keep generated images close to the real ones, in addition to the adversarial loss. Experimental results demonstrate that the proposed approach produces high-quality color image and depth image from low-quality image pair, and it is superior to several other leading methods. Besides, the applications of the proposed method in other tasks are image smoothing and edge detection at the same time.
\end{abstract}


%
\IEEEpeerreviewmaketitle

\section{Introduction}
%
%
%
%
\IEEEPARstart{L}{ow-resolution}  and noisy images are always annoying for a variety of practical applications such as image and video display, surveillance, to name a few. In order to enlarge image's resolution and enhance the quality of super-resolution image, a tremendous amount of works have been developed in the field of color super-resolution (SR) for several decades \cite{Park1, Yang2}. Recently several convolutional neural network (CNN) based methods such as \cite{Dong4, Shi5, Kim6} have reported better super-resolution results than previous methods, whose complexity could be an order of magnitude lower.

One of the earliest CNN-based super-resolution works is three SRCNN in \cite{Dong3}. Latter, the deconvolution operation is used in \cite{Dong4} to directly learn the projection from low resolution (LR) image to high-resolution (HR) image. In \cite{Shi5}, an efficient sub-pixel convolution layer is introduced to learn a series of filters to project the final LR feature maps into HR image. Different from the shallow neural network in \cite{Dong3, Dong4, Shi5}, a very deep convolutional network is presented in \cite{Kim6} to learn image's residuals with extremely high learning rates. However, these methods' objective functions are always the mean squared SR errors, so their SR output images usually fail to have high-frequency details when the up-sampling factor is large. In \cite{Ledig7, Johnson8}, a generative adversarial network is proposed to infer photo-realistic images in terms of the perceptual loss. In addition to the single image SR, image SR with its neighboring viewpoint's high/low resolution image has also been explored. For instance, in \cite{Garcia9} high-frequency information from the neighboring full-resolution views and corresponding depth image are used to enhance the low-resolution view images. In \cite{Jin12}, except mixed resolutions, the multiple LR stereo observations are leveraged to increase image's resolution.

Due to depth information's facilities to many real-world applications, depth SR problems have been widely explored in recent years. When only LR depth image is given, this SR problem is called single depth super-resolution. But, if the LR depth image is available accompanied with HR color image, researchers often name this kind problem of SR after joint depth SR/color image-guided SR. In \cite{Mac13}, by searching a list of HR candidate patches from the database to match with the LR patches, the problem of depth SR is transformed into Markov random field (MRF) labeling problem to reconstruct the full HR image. After that, single depth SR is decomposed as two-step procedures: first the HR edge map is synthesized with HR patches according to the MRF optimization problem; and then a modified joint bilateral filtering is employed to achieve image up-sampling with this HR edge map \cite{Xie14}.

Since the HR color image can be easily got by the consumer camera sensors in most cases, so the available color image can be used as an available prior information to upscaling the LR depth image, under the assumption of structural similarity between color image and depth image. Here, we just classify joint depth SR approaches into three classes: filtering-based methods, optimization methods and CNN-based SR methods. For example, bilateral filtering and guided image filtering are often used to get the interpolation weights to resolve the problem of depth SR \cite{Kopf16, Yang17, Chan18}. The joint bilateral filtering in \cite{Kopf16} use color image as a prior to guide the up-sampling from LR to HR. Meanwhile, bilateral filtering is iteratively used to refine the input low-resolution depth image in \cite{Yang17} to improve the spatial resolution and depth precision. Later, to prevent texture-copy artifacts from color image and against the inherent noisy nature of real-time depth data, an adaptive multi-lateral up-sampling filter in \cite{Chan18} is described to up-sample depth information. In \cite{He19}, a more advanced filtering is called guided filtering, whose ambition is to transfer the structures from a guidance image into a target image.

The second class of joint depth super-resolution methods often build their model by converting SR problems into the convex and non-convex optimization with different prior knowledge to regularize the objective function. For example, a MRF-based model \cite{Diebel20}, which consists of data term and smoothness prior term, is built up to align the discontinuities of depth image with color image's boundaries. However, this model always suffers from the texture-copy artifacts and depth bleeding artifacts, when color image could not provide enough information for depth image reconstruction. Thus, to sharpen depth boundaries and to prevent depth bleeding, a nonlocal means term is incorporated into the MRF model to help local structure to be preserved \cite{Park21}. To suppress texture-copy artifacts and reduce computational cost, variable bandwidth weighting scheme \cite{Liu22} is used into the MRF model to adjust the guidance weights based on depth local smoothness. These methods of \cite{Park21, Liu22} implicitly put the inconsistency between the depth image and the color image into the smoothness term of MRF model. Later, a unified framework proposes to cast guided interpolation into a global weighted least squares optimization framework \cite{Li24}. In \cite{Ferstl25}, the higher order regularization is used to formulate depth image up-sampling as a convex optimization problem. In \cite{Ham29}, a static and dynamic filter (SDF) is designed to address the problem of guided image filtering by jointly using structural information from the guidance image and input image.

Although these recent advanced techniques achieve some appealing performances, they are built on the complex optimization algorithms using hand-designed objective functions, which always have high complexity of computation and limit their widely practical applications. Recently, deep joint image filtering framework based on convolutional neural network is proposed in \cite{Li30} to adaptively transfer co-occurrence information from the guidance image to the target images. Meanwhile, in order to adaptively up-sample depth image's small-scale and large-scale structures, a multi-scale guided convolutional network is trained in high-frequency domain for up-sampling depth map \cite{Hui31}.

Some of literatures always claim that LR depth image can be available and accompanied with HR color image in the dynamic scene, so the majority of these works put their emphasis on HR color image aided depth super-resolution. But they often lose sight of the significance of simultaneously depth image and color image SR with deep learning. As a matter of fact, this task is very important for several 3D video application fields. For example, the 3D-HEVC \cite{Tech32} has leveraged the full-resolution color video and depth video with multi-view video plus depth (MVD) format to compress 3D video. If the techniques of simultaneous depth and color SR can be put into the 3D-HEVC framework, apparently their coding efficiency can be greatly improved. From our investigation, we find some works such as \cite{Li33} have embedded the CNN-based SR into HEVC coder to achieve significant bits saving, so the research of simultaneously depth and color image SR is a meaningful topic for both industry and academia.

Recently, generative adversarial networks \cite{Goodfellow37} is used to generate high-quality image to achieve the tasks of super-resolution and image style transfer, and image-to-image transfer \cite{Johnson8}. In \cite{Johnson8}, perceptual loss function is applied on the tasks of image transformation such as image style transfer by training feed-forward networks. In \cite{Isola36}, a general solution to the problem of image-to-image translation is proposed to finish a lot of tasks, such as synthesizing a new image from the label map, reconstructing a scene image from an edge map.

Following the works of \cite{Johnson8, Isola36}, we propose to use color-depth conditional generative adversarial network (CDcGAN) to deal with both challenging tasks of color SR and depth SR at the same time. Our generative network consists of five components: color feature extraction subnetwork, depth feature extraction subnetwork, color-depth feature merge subnetwork, color image reconstruction subnetwork and depth image reconstruction subnetwork. First, we respectively extract color feature and depth feature in the first two subnetworks and then these features are merged by color-depth feature merge subnetwork, which is inspired by the literature \cite{Li30}. After that, color feature and/or depth feature feed into the last two subnetwork in addition to the merged depth-color features in order to produce HR color-depth images at the same time. Secondly, one discriminator is used to distinguish real color image from the generated color image. The reasons of why depth image SR without discriminator comes from a fact that depth image is not used to directly displayed on the screen, but it is always used as scene's geometry information to direct the rendering of virtual images with each pixel of depth image representing the distance between camera and object. Thus, only three auxiliary losses are taken to regularize depth image SR. Thirdly, in our generative network, three additional loss: data loss, Total Variation (TV) loss, and 8-connected gradient difference loss are also used for color image, so as to ensure that image pairs produced by the generator are similar to the true image pairs.

The rest of paper is organized as follows. First, our approach is presented in Section 2. After experimental results are evaluated in Section 3, we draw a conclusion in Section 4.
\section{Method}

\subsection{Networks}
\begin{figure*}[t]
\centering
\includegraphics[width=7in]{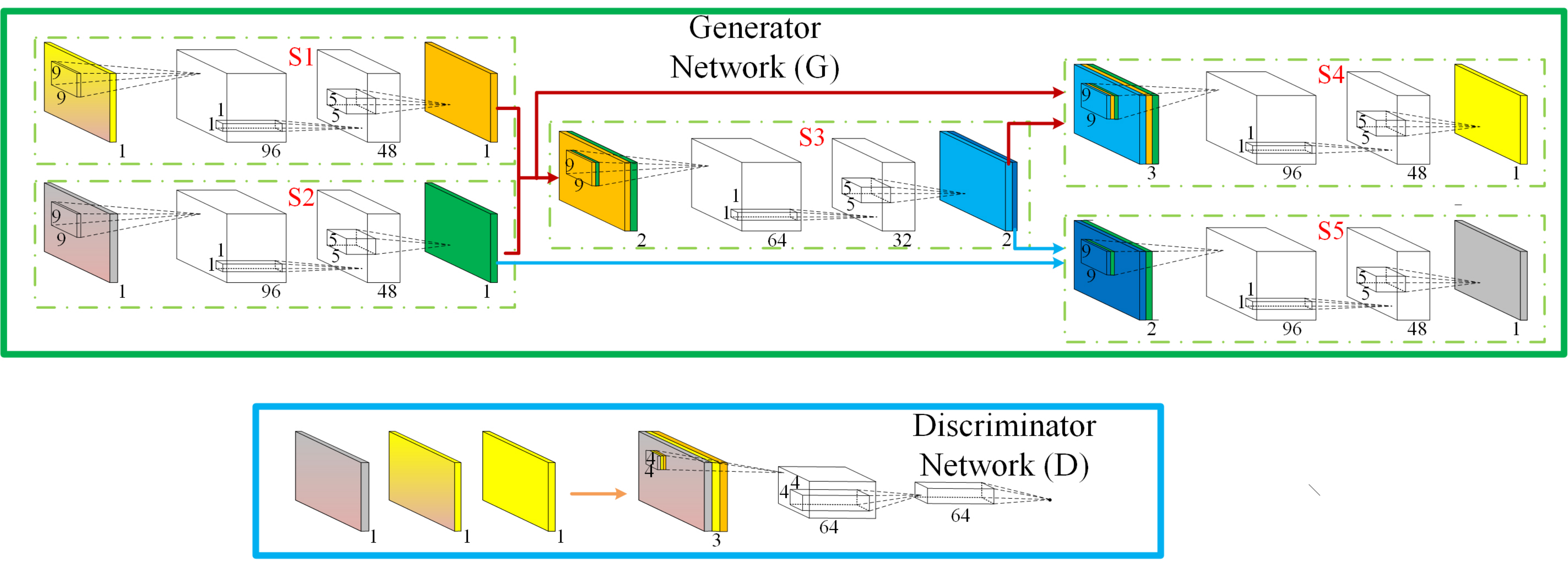}
\caption{\small The diagram of color-depth conditional generative adversarial network(CDcGAN)}
\label{fig::network}
\vspace{-2mm}
\end{figure*}
\begin{figure}[t]
\centering
\includegraphics[width=3in]{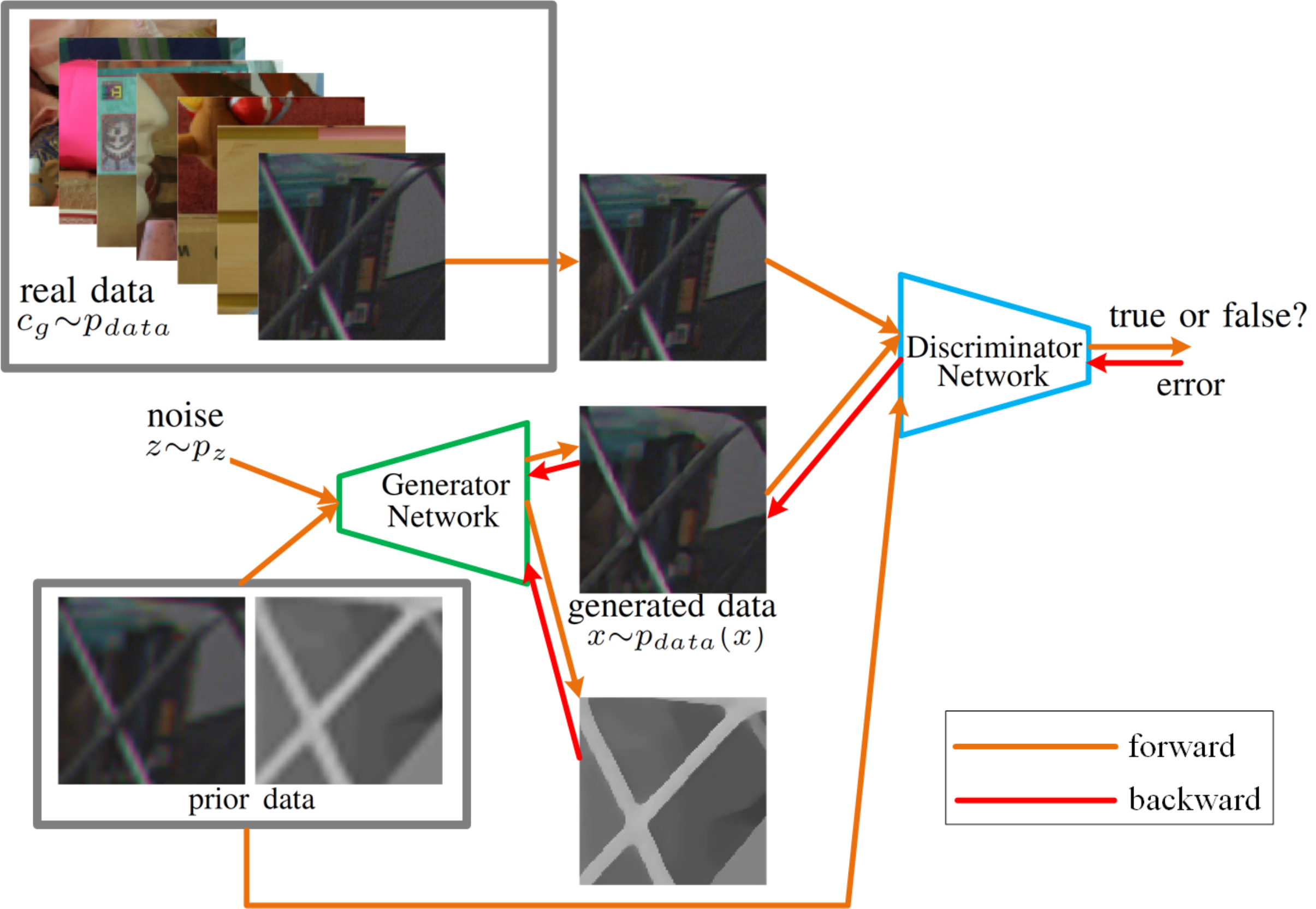}
\caption{\small The workflow of color-depth conditional generative adversarial network}
\label{fig::networkflow}
\vspace{-2mm}
\end{figure}
Given the LR color-depth image pair $(c,d)$, we propose to use the conditional GAN to generate the HR color-depth image pair $(x,y)$. To the best of our knowledge, this is the first deep learning-based color-depth super-resolution scheme. As illustrated in Fig. ~\ref{fig::network} and Fig. ~\ref{fig::networkflow}, our conditional generative adversarial network consists of a generator network (G) and a discriminator network(D). First, our proposed generative network respectively feeds LR depth image $d$ and LR color image $c$ into color feature extraction subnetwork (S1) and depth feature extraction subnetwork (S2) to extract their features, as displayed in Fig. ~\ref{fig::network}. In addition, the depth feature and color feature are fed into color-depth feature merge subnetwork (S3). Finally, the first two subnetworks features and color-depth merged features are leveraged to reconstruct HR color image and  depth image with color reconstruction subnetwork (S4) and depth reconstruction subnetwork (S5) respectively. In particular, the generator $G$ has two subnetworks $S4$ and $S5$ to produce image pair $(x,y)$ from the given image pair $(c,d)$. In the reconstruction subnetwork $S4$, the depth feature maps from the merged subnetwork, color feature extraction layer and depth feature extraction layer are chosen to generate HR color image. However, HR depth images are convolved only with the first feature extraction subnetwork for depth image, in addition to the feature map of the merged subnetwork. In other words, the skip-connection is chosen for both color and depth SR, but for depth image's reconstruction, one skip-connection is used in order to make the depth features only affected by depth features and mutual features shared by color image and depth image. In \cite{Long}, the skip-connection has been successfully used for a semantic segmentation network. Here, we share a similar idea about skip-connection for color-depth super-resolution.

The generated color image is used to fool the discriminator $D$. The discriminator is trained to identify the false images from the true image pair $(c_g,d_g)$ with size of $M \times N$. Note that only one discriminator is used in our generative adversarial network. Actually, we can use two discriminators to distinguish the true image pairs from the false ones. However, the depth image is not watched directly and keeping the accuracy of the depth image is the major task for depth super-resolution, so there is no adversarial loss for depth reconstruction subnetwork (S5).

In each subnetwork, we use three convolutional layers similar to \cite{Dong3}. The advantage of this network lies in the middle convolutional network with kernel size of 1x1 in spatial domain so that the parameters of the networks could greatly decreased, while guaranteeing the nonlinear of this neural network. The S1/S2 three convolutional layers are presented as follows: 9x9x1x96; 1x1x96x48; and 5x5x48x1 respectively. The S3 convolutional layers are is listed as: 9x9x2x64; 1x1x64x32; and 5x5x64x2. In addition, the S4/S5 convolutional layers are 9x9x3x9; 1x1x96x48 and 5x5x48x1. The convolutional layers of S1, S2, and S3 are conducted with stride of 1 and padding of 1, but the ones of S4 and S5 are processed without padding in order to keep the output image's size same as the ground truth image. All the convolutional layers are followed by the activation function of ReLU, except each last convolutional layer of S1, S2, S3, S4, and S5.

Except the SRCNN network \cite{Dong3}, there are many other choices for each component of our generative network. For example, modified VGG deep networks \cite{Simonyan41}, can be used for the first three subnetworks. In \cite{Zhang42}, all the convolution kernel size in spatial domain is 3x3 for image de-noising with modified VGG deep networks. In addition, two reconstruction subnetwork of our network can alternatively choose the deconvolution neural networks or sub-pixel convolutional neural network to reduce the parameter of networks and corresponding computational costs \cite{Dong4, Shi5}.

As depicted in Fig. ~\ref{fig::network}, the discriminator is a three-layer convolutional neural network. The parameters are respectively 4x4x3x64 with stride=2, 4x4x64x64 with stride=2, and 5x5x64x1 with stride=1. In the discriminator network, the first two convolution layers are followed by Leaky ReLU activation function, while the last layer is activated by sigmoid function. There are some alternative networks for our discriminators, such as the Encoder-decoder network or U-net used in \cite{Isola36}.

\subsection{Objective}

In our objective function, there is no adversarial loss for the depth image. Instead, three auxiliary losses are considered to make the generated depth image close to the truth image. Contrary to depth images, which only contain sharp boundaries and some flat or piece-wise smooth regions, color images usually have more informative textural details. So it is important for color images to be more realistic compared to the true image, especially when the up-sampling factor is very large.

In a summary, the objective of our model can be expressed as follows:
\begin{equation}
\begin{split}
G^*= \min_{G} \max_{D} \alpha \cdot L_{CDcGAN}(G,D)\\+ (L_{data}(G)+L_{TV}(G)+L_{GD}(G)),
 \label{eqn::cdgan_problem}
 \end{split}
\end{equation}
where $L_{CDcGAN}(G,D)$ is the adversarial loss and the others are three auxiliary losses. They will be defined later. Here, the parameter $\alpha$ is used to adjust the contribution for color-depth super-resolution between the GAN loss and three auxiliary losses.

\subsection{Adversarial Loss}
For brevity of latter description, we denote the true color image data's distribution and generated color image data's distribution as $p_{data}(c_g)$ and $p_{data}(x)$, while $p_{z}$ is input noise's distribution. As shown in Fig. ~\ref{fig::network}, the generator $G(c,d,z)$ is used as a mapping from the LR image set $(c,d)$ to HR image one $(c_g,d_g)$. $D(c,d,c_g)$ describes the probability that $c_g$ comes from the true image data rather than image data produced by the generator $G(c,d,z)$, while the probability of data from $G(c,d,z)$ is represented as $D(c,d,G(c,d,z))$.

In our model, the adversarial loss is expressed as follows:
\begin{equation}
\begin{split}
L_{CDcGAN}(G,D)= E_{c_g \sim p_{data}(c_g)} \log(D(c,d,c_g)) +\\E_{x \sim p_{data}(x),z \sim p_{z}}
[\log(1-D(c,d,G(c,d,z)))]
\end{split}
 \label{eqn::GANs}
\end{equation}
in which z is random noise.

\subsection{Auxiliary Losses}
In our objective function, three auxiliary losses are included: data loss, TV loss, and 8-connected gradient difference loss, which are leveraged to make image pair $(x,y)$ produced by the generator G to be similar enough to the true image pair $(c_g,d_g)$. The vectors of $(x,y)$ and $(c_g,d_g)$ are represented as $\bm{(X,Y)}$, and  $\bm{(C_g,D_g)}$. Like the traditional TV model, the data loss keeps the output's value consistent to the ground truth value, while the TV loss $L_{TV}(G)$ emphasizes the correlation of the output's values with its neighboring pixel's value in order to keep generated image to be smooth and robust against noises. Our data loss function $L_{data}(G)$, including both color image's data loss and depth image's data loss, is defined as follows:
\begin{equation}
\begin{split}
L_{data}(G)= \frac{1}{M \cdot N}\sum_{i}(||\bm{X}(i)-\bm{C}_g(i)||_L\\ + ||\bm{Y(i)}-\bm{D}_g(i)||_L),
\end{split}
 \label{eqn::data loss}
\end{equation}
where $||\cdot||_L$ represents the L norm.

Our TV loss function is defined as follows:
\begin{equation}
\begin{split}
L_{TV}(G)=   \frac{1}{M \cdot N}\sum_{i}((|| \nabla_x \bm{X}(i)||_L +\\ ||\nabla_y \bm{X}(i)||_L) +  (|| \nabla_x \bm{Y}(i)||_L + || \nabla_y \bm{Y}(i)||_L))
\end{split}
 \label{eqn::TV loss}
\end{equation}
where $\nabla_x$, and $\nabla_y$ are the gradients in the x-direction and y-direction.

Here, we use the 8-neighboring gradient difference (GD) loss to make generated image pair's gradient information to be similar to that of the ground truth image in the gradient domain. The 8-neighboring GD loss is defined as follows:
\begin{equation}
\begin{split}
L_{GD}(G)= \frac{1}{M \cdot N} \sum_{i} ((\sum_{k\in{\Omega}}||\nabla_k \bm{X}(i))-\nabla_k \bm{C}_g(i)||_L) +\\  (\sum_{k\in{\Omega}}||\nabla_k \bm{Y}(i)-\nabla_k \bm{(D}_g(i))||_L))
\end{split}
\label{eqn::GD loss}
\end{equation}
where $\Omega$ is each pixel's neighbourhood, and $\nabla_k$ is the $k$-th gradient between each pixel and $k$-th pixels among 8-neighbouring pixels.

In some literatures \cite{Mathieu39}, it has been reported that L2 loss always leads to image blurring. In \cite{Hui31, Isola36}, the traditional loss, such as L1 distance, has been added into the GAN objective function, in which the generator aims to not only fool the discriminator but also to make sure that generated samples move towards the ground truth in an L1 sense. Thus, in the proposed neural network, the L1 loss is used in our three auxiliary losses to keep generated sample close enough to the real one, and make images to be sharp rather than blurring.

\subsection{Other Applications}
Our proposed network is not restricted to finish the task of depth-color super-resolution. In fact, similar networks can be specifically designed for different tasks, e. g. simultaneous edge detection and semantic segmentation, concurrent edge detection and image smoothing, and even finishing these three tasks at the same time, when corresponding networks have one input (e.g., color image) or two inputs (e.g. color image and depth image).

For image smoothing and edge detection at the same time, we change the output feature numbers of each subnetwork's last convolutional layer. For example, two inputs in our network respectively have three channels and six channels, then the last convolutional layer output feature map numbers in the the S1 and S2 / in S4 and S5 will be 3 and 6 respectively. Here, one input image is color image, while another one is composed of the six gradient map of color image in both vertical and horizonal direction. Note that learning image smoothing use the content loss, TV loss, and gradient loss, but learning edge detection only employs content loss. And each convolution layer is padded to be the same size as the input features.

In fact, it can be extended into multiple inputs and multiple outputs with one network. In many cases, several images of a scene with different modalities or different lighting conditions are observed at the same time, so one or more images in other modalities are desire to be generated, when some modal images are known.

\section{Experimetal results and analysis}
To validate the efficiency of the proposed architecture for super-resolution, we compare our method with the Bicubic method, SRCNN \cite{Dong3}, and VDSR \cite{Kim6} for color image SR. In addition, for depth super-resolution, not only the results of single depth SR with SRCNN \cite{Dong3}, and VDSR \cite{Kim6} are given, we also compare joint depth super-resolution results with several existing methods such as GIF \cite{He19}, FGS \cite{Li24}, RGIF \cite{Ham29}, TGV \cite{Ferstl25}, RGDR \cite{Liu22}, HQDU \cite{Ferstl25}, MRF \cite{Diebel20}. Three measurements of image quality, e.g. Peak Signal-to-Noise Ratio (PSNR), Structural SIMilarity (SSIM) index, and image sharpness \cite{Mathieu39}, are used for the comparison of different methods. Finally, we also use our architecture to learn filters for simultaneously image smoothing and edge detection.

\subsection{Implementation details}
Our architecture of simultaneously color-depth super-resolution is implemented in TensorFlow \cite{Abadi43} including about 200 thousand parameters, but the generator only use 92358 parameters. We train our neural network with 100,000 image color-depth patches with size $32\times32$ from 90 color-depth image pairs. In our training dataset, 52 color-depth image pairs come from the Middlebury dataset, and the remaining color-depth image pairs are got from the MPI Sintel color-depth dataset. In our model, $\alpha$ equal to 0.002. We train our model for 30 epochs using Adam, where the beta1=0.5, beta2=0.999, and the learning rate is set to be 0.0002. Note that the hyper-parameters of beta1 and beta2 of Adam control the exponential decay rates of moving averages, whose details can be found in \cite{Kingma46}. During training, the the parameters of the discriminator $D$ are updated by Adam, which is followed by the updating of generator's ones. After alternative training the generator $G$ and discriminator $D$ up to Nash equilibrium, the generator $G$ becomes powerful to produce high-quality image, as shown in Fig. ~\ref{fig::networkflow}.

In order to further validate the efficiency of our architecture, we use the same architecture in Fig. ~\ref{fig::network} to learn filters for image smoothing and edge detection at the same time. we use the BSDS500 dataset from Berkeley Computer Vision Group and corresponding smoothed image with L0 gradient minimization in \cite{Xus} as our training data for image smoothing and edge detection. we augment data for training by rotating image. Specifically, 100,000 patches with size $64\times64$ are extracted from these augmented data. The other training parameters are set the same as the ones described above in simultaneously color and depth super-resolution.
\begin{table*}[thb!]
\small
\centering
{
\caption{The objective comparison of color image super-resolution for 2x and 4x up-sampling factor}
\label{tbl::color-objectivemeasure}
\scriptsize
\begin{tabular}{|c|c|c|c|c|c|c|c|c|c|}
\hline
\multicolumn{ 1}{|c}{Seq} & \multicolumn{ 1}{|c|}{Bicubic} &      SRCNN &       VDSR & \multicolumn{ 1}{c|}{CDcGAN} & \multicolumn{ 1}{c}{Seq} & \multicolumn{ 1}{|c|}{Bicubic} &      SRCNN &       VDSR & \multicolumn{ 1}{c|}{CDcGAN} \\

\multicolumn{ 1}{|c}{} & \multicolumn{ 1}{|c|}{} &        \cite{Dong3} &        \cite{Kim6} & \multicolumn{ 1}{c|}{} & \multicolumn{ 1}{c}{} & \multicolumn{ 1}{|c|}{} &        \cite{Dong3} &        \cite{Kim6} & \multicolumn{ 1}{c|}{} \\
\hline
                                      \multicolumn{ 5}{|c|}{2x} &                                       \multicolumn{ 5}{c}{4x} \\
\hline
                                                                                                    \multicolumn{ 10}{|c}{M1} \\
\hline
         B &       38.5 &      40.58 &      40.95 &      40.05 &          B &      32.71 &      34.11 &      34.16 &      34.49 \\
\hline
         L &      39.21 &      38.79 &      40.28 &      39.19 &          L &      33.31 &      28.75 &      34.56 &      34.01 \\
\hline
         N &      40.66 &      38.32 &      42.98 &      41.22 &          N &      33.35 &      25.54 &       35.3 &      34.86 \\
\hline
         U &      33.77 &      34.04 &      36.04 &      35.55 &          U &      29.89 &       26.10 &      31.47 &      31.49 \\
\hline
         S &      38.89 &      38.87 &      41.51 &      39.93 &          S &      33.64 &      32.55 &      34.43 &      34.44 \\
\hline
{\bf Ave.} &      38.21 &      38.12 & {\bf 40.35} &      39.19 & {\bf Ave.} &      32.58 &      29.41 & {\bf 33.98} &      33.86 \\
\hline
                                                                                                    \multicolumn{ 10}{|c}{M2} \\
\hline
         B &      0.926 &      0.934 &      0.936 &      0.928 &          B &      0.833 &      0.849 &      0.854 &       0.86 \\
\hline
         L &       0.97 &      0.967 &      0.974 &      0.969 &          L &      0.882 &       0.85 &      0.894 &      0.891 \\
\hline
         N &      0.968 &      0.965 &      0.973 &       0.97 &          N &      0.894 &      0.869 &      0.909 &      0.906 \\
\hline
         U &      0.881 &       0.89 &      0.912 &      0.906 &          U &      0.732 &      0.729 &      0.756 &       0.77 \\
\hline
         S &      0.958 &      0.958 &      0.967 &      0.967 &          S &      0.875 &      0.859 &      0.885 &      0.889 \\
\hline
{\bf Ave.} &      0.941 &      0.943 & {\bf 0.952} &      0.948 & {\bf Ave.} &      0.843 &      0.831 &      0.859 & {\bf 0.863} \\
\hline
                                                                                               \multicolumn{ 10}{|c}{M3} \\
\hline
         B &      45.41 &      45.98 &      46.17 &      45.88 &          B &      43.53 &      43.77 &      43.91 &      43.93 \\
\hline
         L &      46.82 &      46.55 &      47.05 &      46.72 &          L &      44.23 &      43.23 &      44.53 &      43.98 \\
\hline
         N &      46.82 &      46.65 &      47.51 &      47.12 &          N &      44.09 &      43.59 &      44.59 &      44.24 \\
\hline
         U &      43.55 &      43.69 &      44.39 &      44.21 &          U &      41.95 &      41.82 &      42.44 &      42.38 \\
\hline
         S &      46.48 &      46.38 &      47.19 &      47.04 &          S &      44.55 &      43.93 &      44.73 &      44.54 \\
\hline
{\bf Ave.} &      45.82 &      45.85 & {\bf 46.46} &      46.19 & {\bf Ave.} &      43.67 &      43.27 & {\bf 44.04} &      43.81 \\
\hline
\end{tabular}

}
\end{table*}

\begin{table*}[thb!]
\small
\centering
{
\caption{The objective comparison of depth super-resolution for 2x up-sampling factor}
\label{tbl::depth-objectivemeasureX2}
\scriptsize
\begin{tabular}{|c|c|c|c|c|c|c|c|c|c|c|c|c|}
\hline
\multicolumn{ 1}{|c|}{SEQ} & \multicolumn{ 1}{c|}{M} & \multicolumn{ 1}{c|}{Bicubic} &      SRCNN &       VDSR &        GIF &        FGS &       RGIF &        TGV &       RGDR &       HQDU &        MRF &        CDc \\

\multicolumn{ 1}{|c|}{} & \multicolumn{ 1}{c|}{} & \multicolumn{ 1}{c|}{} &        \cite{Dong3} &        \cite{Kim6} &       \cite{He19} &       \cite{Li24} &       \cite{Ham29} &       \cite{Ferstl25} &       \cite{Liu22} &       \cite{Park21} &       \cite{Diebel20} &      GAN \\
\hline
         B & \multicolumn{ 1}{c|}{} &     41.53  &     44.37  &     46.88  &     32.91  &     34.85  &     39.54  &     38.37  &     36.48  &     37.74  &     36.70  &     46.35  \\

         L & \multicolumn{ 1}{c|}{} &     48.97  &     50.34  &     52.86  &     41.08  &     43.49  &     47.13  &     46.09  &     42.34  &     46.24  &     44.92  &     54.07  \\

         N & \multicolumn{ 1}{c|}{1} &     43.12  &     46.12  &     48.49  &     33.44  &     35.90  &     40.52  &     38.91  &     37.31  &     38.74  &     37.61  &     47.13  \\

         U & \multicolumn{ 1}{c|}{} &     45.88  &     49.37  &     52.51  &     45.88  &     45.42  &     45.72  &     44.26  &     46.13  &     43.36  &     43.51  &     50.45  \\

         S & \multicolumn{ 1}{c|}{} &     39.54  &     40.49  &     42.21  &     34.33  &     37.75  &     37.21  &     37.95  &     38.24  &     37.43  &     36.77  &     42.64  \\

{\bf Ave.} & \multicolumn{ 1}{c|}{} &     43.81  &     46.14  & {\bf 48.59 } &     37.53  &     39.48  &     42.02  &     41.11  &     40.10  &     40.70  &     39.90  &     48.13  \\
\hline
         B & \multicolumn{ 1}{c|}{} &     0.980  &     0.984  &     0.990  &     0.885  &     0.906  &     0.952  &     0.968  &     0.907  &     0.958  &     0.944  &     0.990  \\

         L & \multicolumn{ 1}{c|}{} &     0.993  &     0.994  &     0.996  &     0.964  &     0.974  &     0.984  &     0.987  &     0.966  &     0.988  &     0.983  &     0.995  \\

         N & \multicolumn{ 1}{c|}{2} &     0.986  &     0.989  &     0.993  &     0.904  &     0.923  &     0.961  &     0.965  &     0.925  &     0.967  &     0.956  &     0.992  \\

         U & \multicolumn{ 1}{c|}{} &     0.996  &     0.996  &     0.998  &     0.985  &     0.994  &     0.992  &     0.993  &     0.991  &     0.990  &     0.993  &      0.999 \\

         S & \multicolumn{ 1}{c|}{} &     0.967  &     0.965  &     0.977  &     0.924  &     0.949  &     0.944  &     0.955  &     0.935  &     0.940  &     0.946  &     0.970  \\

{\bf Ave.} & \multicolumn{ 1}{c|}{} &     0.985  &     0.986  & {\bf 0.991 } &     0.933  &     0.949  &     0.967  &     0.974  &     0.945  &     0.968  &     0.964  & {\bf 0.989 } \\
\hline
         B & \multicolumn{ 1}{c|}{} &      50.42 &      50.98 &      52.61 &      47.37 &      47.94 &         49 &      49.33 &      49.21 &      49.05 &      48.44 &         53 \\

         L & \multicolumn{ 1}{c|}{} &      55.25 &      55.22 &      56.53 &      52.56 &      53.23 &      53.82 &      54.05 &      53.53 &      54.05 &      53.49 &      56.46 \\

         N & \multicolumn{ 1}{c|}{3} &      51.88 &      52.43 &      53.86 &      48.56 &      49.06 &      50.17 &      50.32 &      50.01 &      50.25 &      49.63 &      53.72 \\

         U & \multicolumn{ 1}{c|}{} &      57.91 &       58.9 &      61.25 &      54.96 &       57.5 &      56.34 &      56.23 &      55.93 &      56.02 &      56.46 &      64.67 \\

         S & \multicolumn{ 1}{c|}{} &      50.98 &      50.72 &      52.23 &       49.8 &      51.35 &      50.28 &      50.69 &      51.48 &      50.79 &      50.43 &      52.97 \\

{\bf Ave.} & \multicolumn{ 1}{c|}{} &      53.29 &      53.65 &       55.30 &      50.65 &      51.82 &      51.92 &      52.12 &      52.03 &      52.03 &      51.69 & {\bf 56.16} \\
\hline
\end{tabular}

}
\end{table*}

\begin{table*}[thb!]
\small
\centering
{
\caption{The objective comparison of depth super-resolution for 4x up-sampling factor}
\label{tbl::depth-objectivemeasureX4}
\scriptsize
\begin{tabular}{|c|c|c|c|c|c|c|c|c|c|c|c|c|}
\hline
\multicolumn{ 1}{|c|}{SEQ} & \multicolumn{ 1}{c|}{M} & \multicolumn{ 1}{c|}{Bicubic} &      SRCNN &       VDSR &        GIF &        FGS &       RGIF &        TGV &       RGDR &       HQDU &        MRF &        CDc \\

\multicolumn{ 1}{|c|}{} & \multicolumn{ 1}{c|}{} & \multicolumn{ 1}{c|}{} &        \cite{Dong3} &        \cite{Kim6} &       \cite{He19} &       \cite{Li24} &       \cite{Ham29} &       \cite{Ferstl25} &       \cite{Liu22} &       \cite{Park21} &       \cite{Diebel20} &      GAN \\
\hline
         B & \multicolumn{ 1}{c|}{} &     37.91  &     39.70  &     42.83  &     32.82  &     31.19  &     37.24  &     35.14  &     34.42  &     33.92  &     33.01  &     41.72  \\

         L & \multicolumn{ 1}{c|}{} &     45.54  &     46.42  &     49.43  &     40.98  &     40.83  &     45.01  &     43.58  &     41.04  &     42.54  &     41.36  &     47.43  \\

         N & \multicolumn{ 1}{c|}{1} &     39.22  &     40.80  &     43.99  &     33.35  &     32.23  &     38.46  &     35.42  &     35.16  &     34.37  &     33.56  &     41.88  \\

         U & \multicolumn{ 1}{c|}{} &     42.33  &     38.59  &     46.09  &     40.82  &     42.77  &     42.55  &     42.67  &     42.68  &     40.35  &     40.07  &     51.23  \\

         S & \multicolumn{ 1}{c|}{} &     36.42  &     27.66  &     37.83  &     34.00  &     35.23  &     35.83  &     35.44  &     35.60  &     34.56  &     33.93  &     37.85  \\

{\bf Ave.} & \multicolumn{ 1}{c|}{} &     40.28  &     38.63  & {\bf 44.03 } &     36.39  &     36.45  &     39.82  &     38.45  &     37.78  &     37.15  &     36.39  & {\bf 44.02 } \\
\hline
         B & \multicolumn{ 1}{c|}{} &     0.951  &     0.957  &     0.975  &     0.883  &     0.871  &     0.937  &     0.933  &     0.890  &     0.913  &     0.900  &     0.962  \\

         L & \multicolumn{ 1}{c|}{} &     0.984  &     0.985  &     0.990  &     0.964  &     0.964  &     0.979  &     0.976  &     0.962  &     0.973  &     0.969  &     0.986  \\

         N & \multicolumn{ 1}{c|}{2} &     0.964  &     0.967  &     0.981  &     0.903  &     0.892  &     0.950  &     0.927  &     0.909  &     0.927  &     0.917  &     0.969  \\

         U & \multicolumn{ 1}{c|}{} &     0.991  &     0.971  &     0.995  &     0.985  &     0.991  &     0.989  &     0.989  &     0.989  &     0.987  &     0.986  &     0.997  \\

         S & \multicolumn{ 1}{c|}{} &     0.943  &     0.657  &     0.952  &     0.920  &     0.928  &     0.935  &     0.938  &     0.923  &     0.928  &     0.918  &     0.952  \\

{\bf Ave.} & \multicolumn{ 1}{c|}{} &     0.967  &     0.908  & {\bf 0.979 } &     0.931  &     0.929  &     0.958  &     0.953  &     0.935  &     0.945  &     0.938  &     0.973  \\
\hline
         B & \multicolumn{ 1}{c|}{} &     48.22  &     48.43  &     50.06  &     47.37  &     47.58  &     48.06  &     47.85  &     48.21  &     47.60  &     47.64  &     49.89  \\

         L & \multicolumn{ 1}{c|}{} &     53.35  &     53.24  &     54.44  &     52.55  &     52.95  &     53.10  &     52.95  &     53.06  &     52.73  &     52.70  &     54.23  \\

         N & \multicolumn{ 1}{c|}{3} &     49.58  &     49.75  &     51.24  &     48.54  &     48.59  &     49.36  &     48.76  &     49.08  &     48.66  &     48.73  &     50.81  \\

         U & \multicolumn{ 1}{c|}{} &     55.30  &     53.74  &     57.57  &     54.89  &     55.94  &     54.92  &     55.34  &     54.97  &     54.33  &     54.50  &     58.19  \\

         S & \multicolumn{ 1}{c|}{} &     49.57  &     49.62  &     50.10  &     49.79  &     50.65  &     49.74  &     50.30  &     50.59  &     49.79  &     49.78  &     50.97  \\

{\bf Ave.} & \multicolumn{ 1}{c|}{} &     51.21  &     50.95  &     52.68  &     50.63  &     52.95  &     51.04  &     51.04  &     51.18  &     50.62  &     50.67  & {\bf 52.82 } \\
\hline
\end{tabular}

}
\end{table*}

\begin{figure}[t]
\centering
\includegraphics[width=3in]{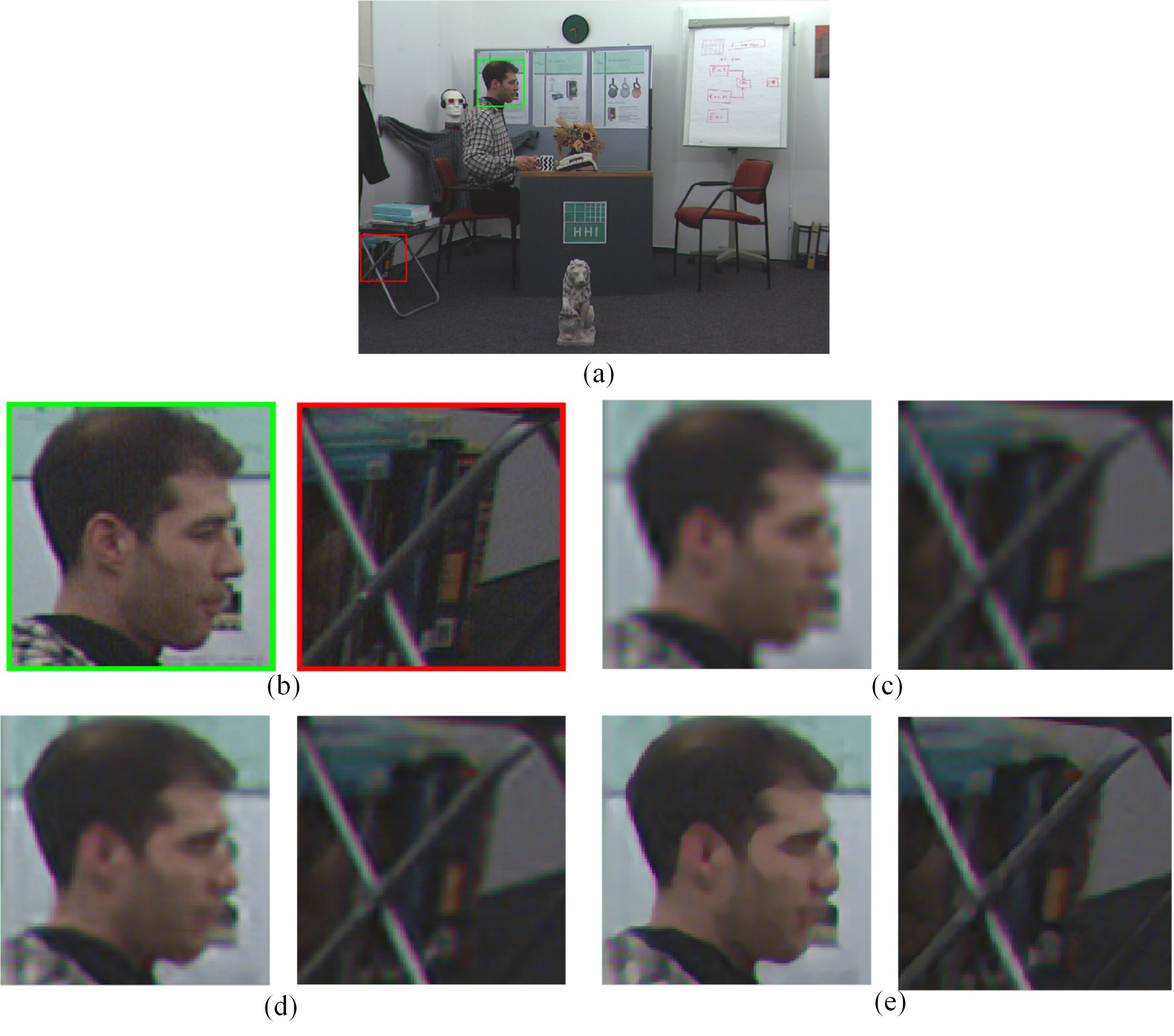}
\caption{\small The SR results for color image with 4x scaling factor. (a) the first frame of Book\_Arrival, (b) the close-ups of (a), (c-f) the close-ups of the results respectively with Bicubic interpolation, SRCNN \cite{Dong3}, VDSR \cite{Kim6}, and our CDcGAN }
\label{fig::bookcolor}
\vspace{-2mm}
\end{figure}

\begin{figure}[t]
\centering
\includegraphics[width=3in]{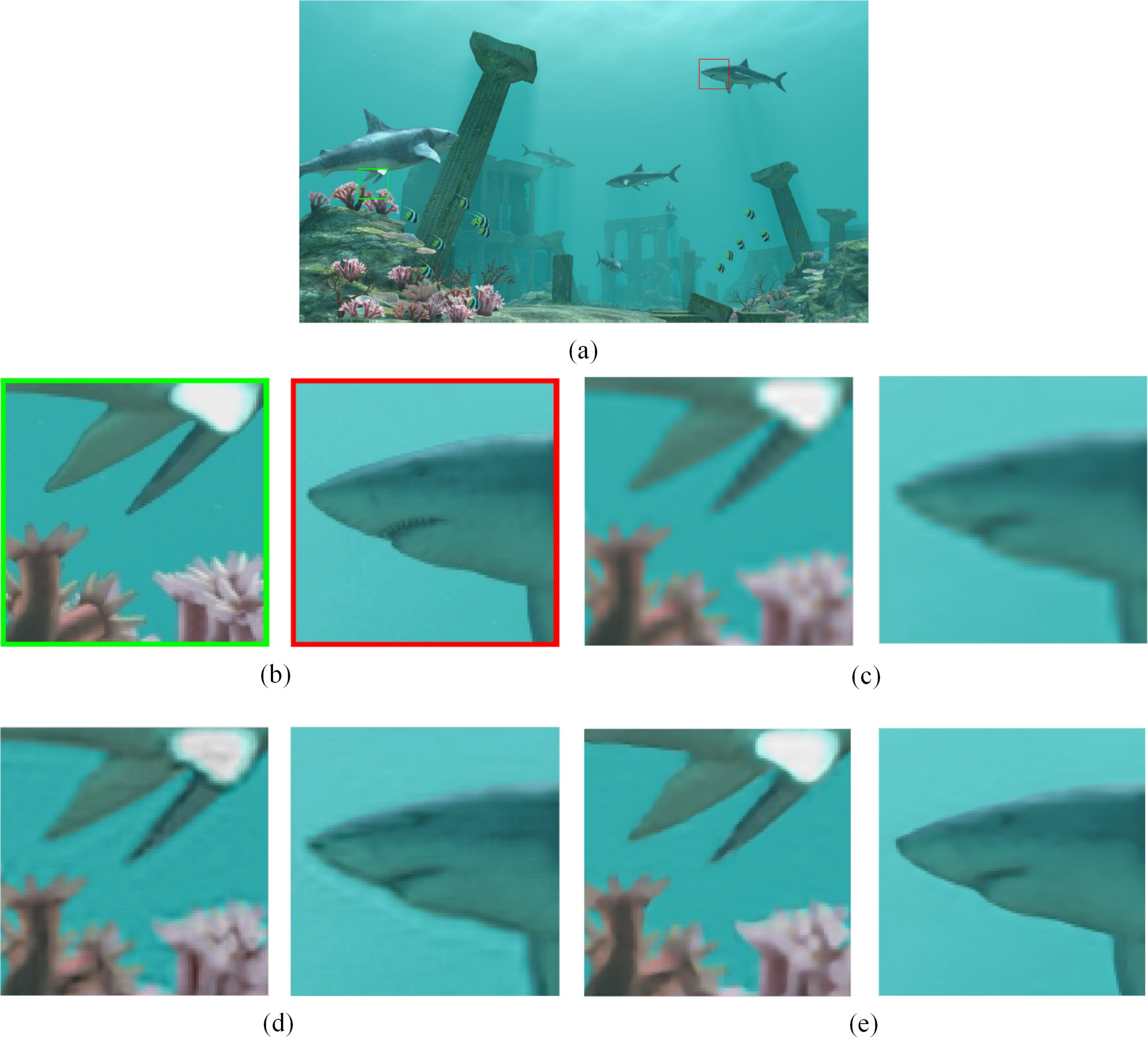}
\caption{\small The SR results for color image with 4x scaling factor. (a) the first frame of Shark, (b) the close-ups of (a), (c-f) the close-ups of the results respectively with Bicubic interpolation, SRCNN \cite{Dong3}, VDSR \cite{Kim6}, and our CDcGAN}
\label{fig::sharkcolor}
\vspace{-2mm}
\end{figure}

\begin{figure}[t]
\centering
\includegraphics[width=3in]{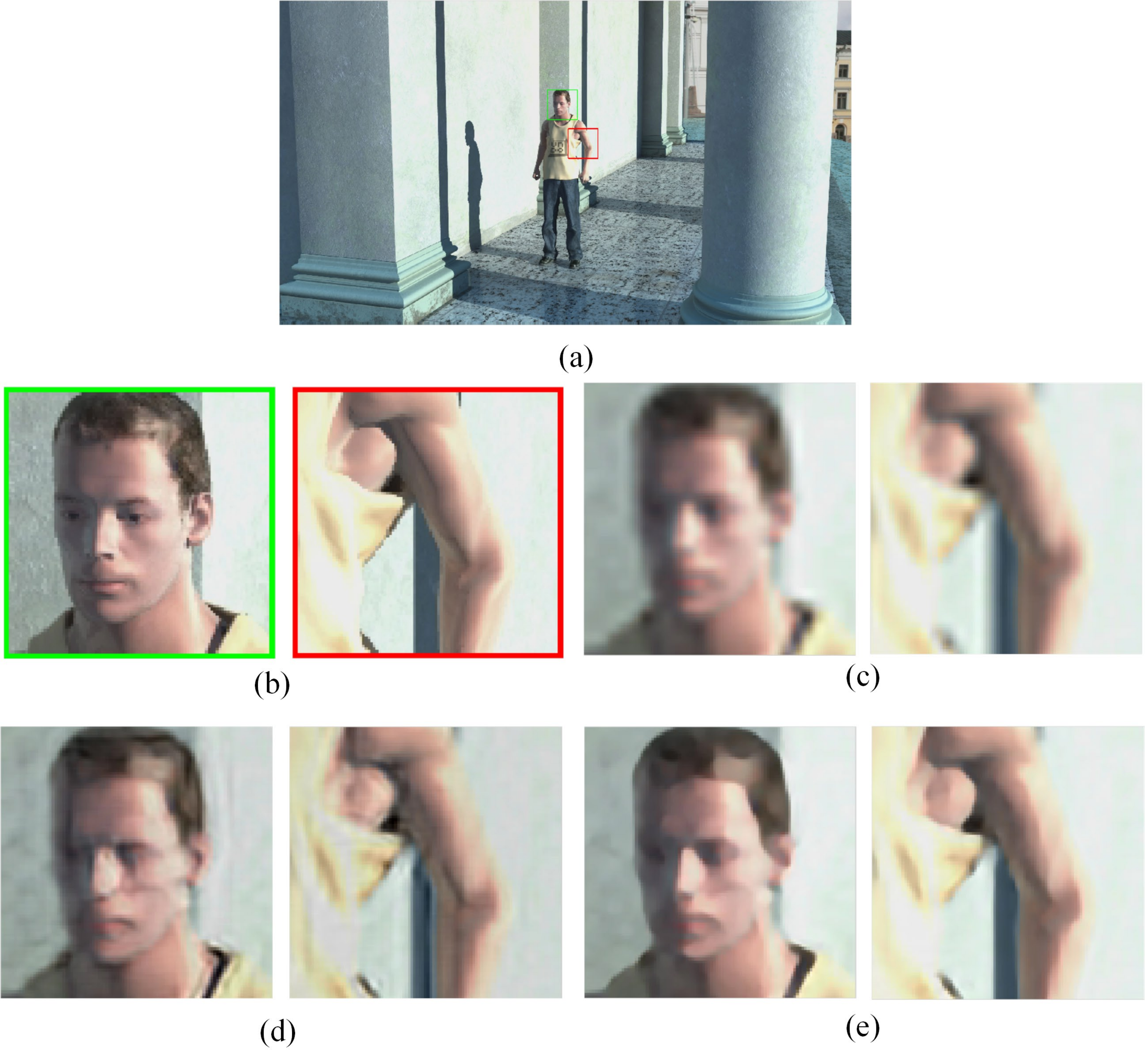}
\caption{\small The SR results for color image with 4x scaling factor. (a) the first frame of Undo\_Dancer, (b) the close-ups of (a), (c-f) the close-ups of the results respectively with Bicubic interpolation, SRCNN \cite{Dong3}, VDSR \cite{Kim6}, and our CDcGAN}
\label{fig::undocolor}
\vspace{-2mm}
\end{figure}

\begin{figure}[t]
\centering
\includegraphics[width=3in]{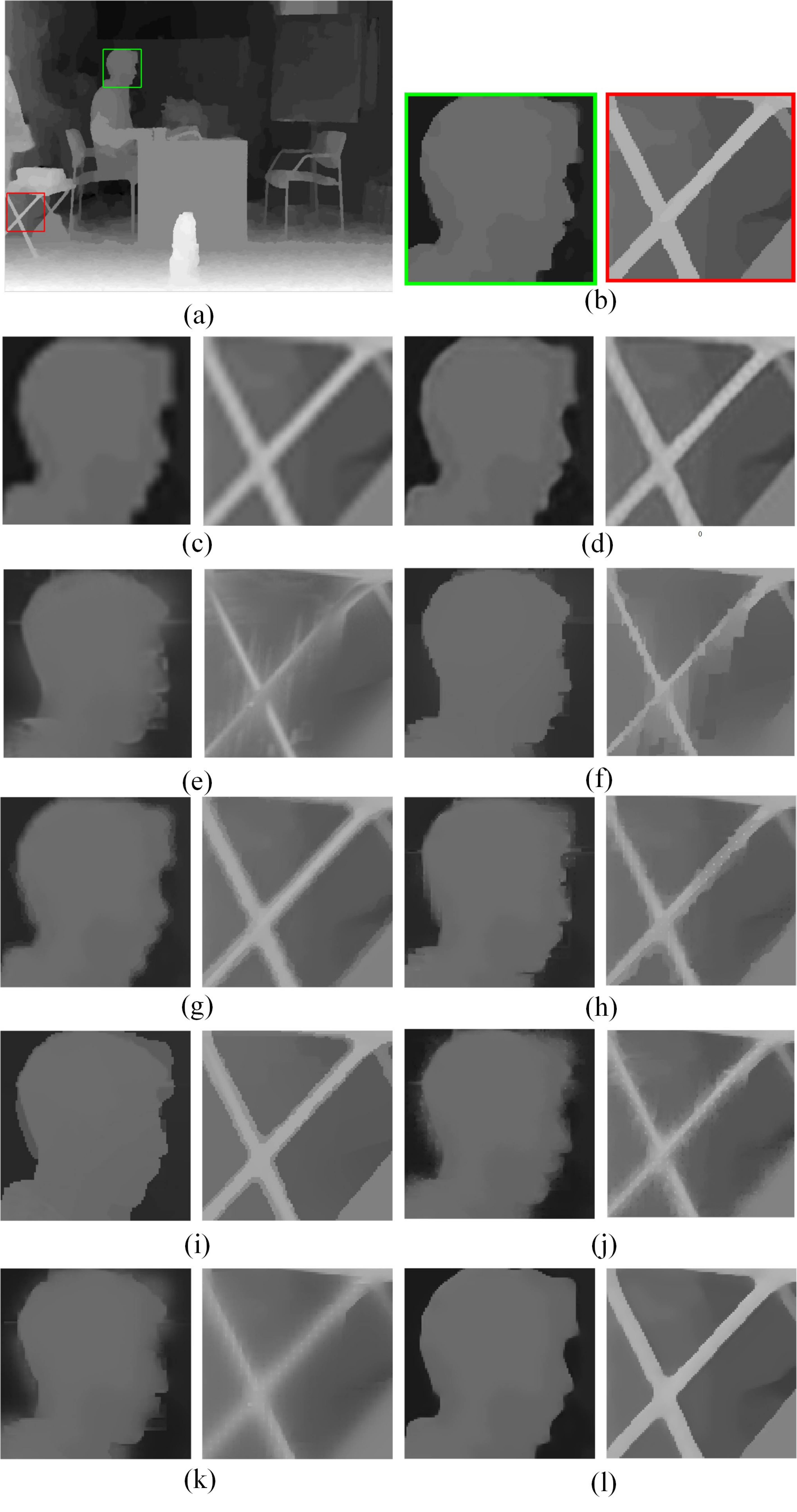}
\caption{\small The SR results for depth image with 4x scaling factor. (a) the first frame of Book\_Arrival, (b) the close-ups of (a), (c-m) the close-ups of the results respectively with Bicubic interpolation, SRCNN \cite{Dong3}, VDSR \cite{Kim6}, GIF \cite{He19}, FGS \cite{Li24}, RGIF \cite{Ham29}, TGV \cite{Ferstl25}, RGDR \cite{Liu22}, HQDU \cite{Park21}, MRF \cite{Diebel20}, and our CDcGAN}
\label{fig::bookdepth}
\vspace{-2mm}
\end{figure}

\subsection{The objective and visual quality comparison for super-resolution}
We use five standard 3D sequences to show the efficiency of the proposed method. The five testing color-depth sequences of first 100 frames contain Love\_Bird (denoted as L), Book\_Arrival (B), and Newspaper (N) with resolution of 768x1024, Shark (S), and Undo\_dancer (U) with size of 1920x1088. Three objective quality of both color SR and depth SR are evaluated in terms of PSNR, SSIM, and sharpness. The comparative results are displayed in Table 1-3, where PSNR, SSIM, and sharpness are respectively denoted as M1, M2, M3.

From the Table ~\ref{tbl::color-objectivemeasure}, it can be seen that both PSNR and sharpness of VDSR \cite{Kim6} are better than ours and other approaches for 2x and 4x color super-resolution, but the SR image with VDSR tends to be blurring to some degree, especially when up-sampling factor is 4, as displayed in Fig. ~\ref{fig::bookcolor}, ~\ref{fig::sharkcolor}, ~\ref{fig::undocolor}. However, for color up-sampling factor of 2x and 4x, the proposed method keeps generated image sharp enough and the visual quality is competitive against than VDSR \cite{Kim6}, in despite of the very deep neural networks used in VDSR \cite{Kim6}. The SSIM measurement of our approach is better than that of SRCNN \cite{Dong3}, but is slight lower than the one of VDSR for 2x super-resolution. Our method's SSIM performs better than SRCNN's \cite{Dong3} and VDSR's \cite{Kim6} for 4x color image super-resolution. In a summary, our method has better visual performance on image reconstruction and is robust to noise, which benefits from that the TV loss ensures the generated color image's flat regions to be smooth, and the gradient difference loss tends to keep the color image similar enough to the ground truth color image in the gradient domain. Thus, the generated color image better obey the real sample's distribution when conditional GAN is used for color super-resolution.

\begin{figure}[ht]
\centering
\includegraphics[width=3in]{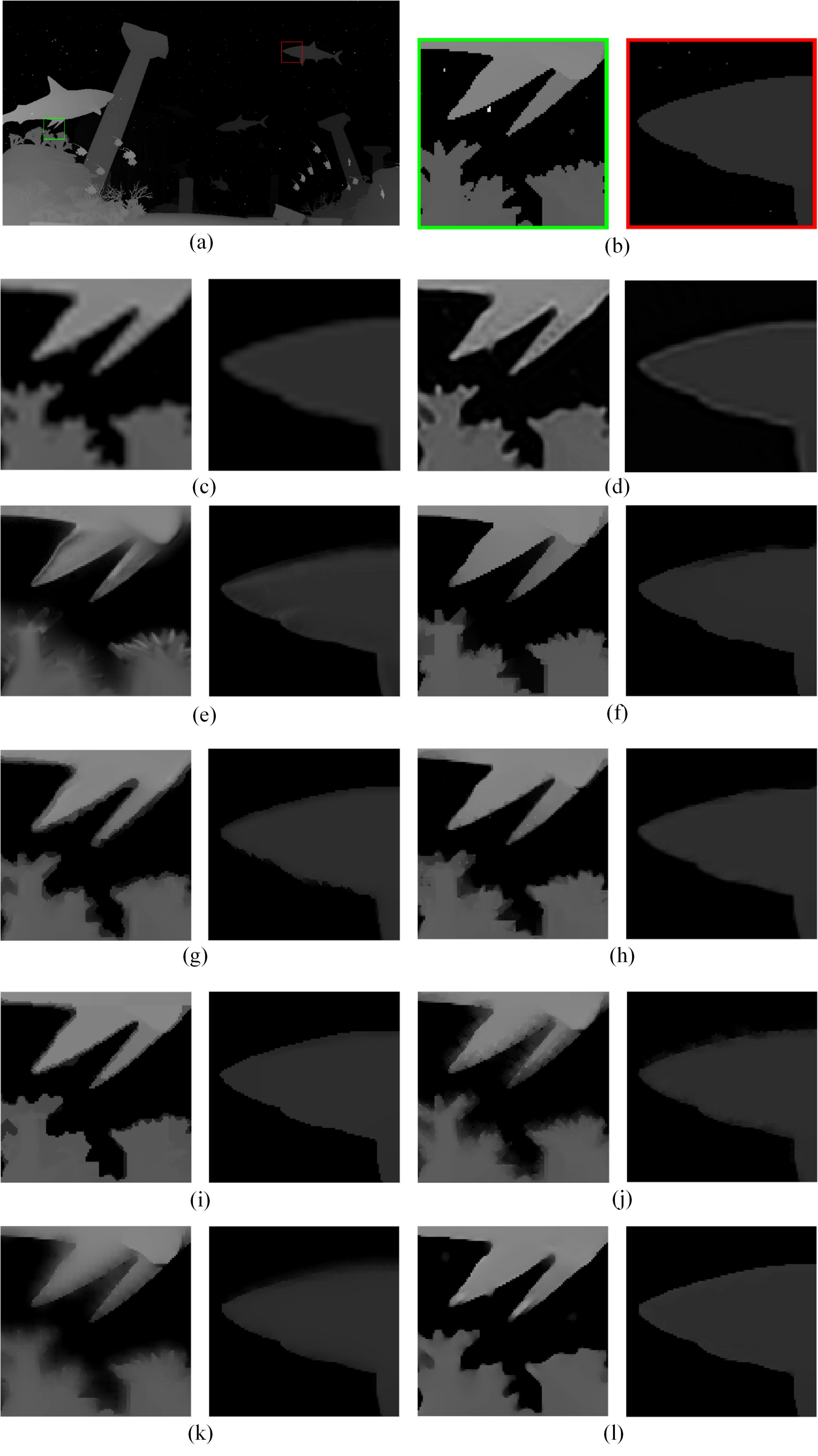}
\caption{\small The SR results for depth image with 4x scaling factor. (a) the first frame of Shark, (b) the close-ups of (a), (c-m) the close-ups of the results respectively with Bicubic interpolation, SRCNN \cite{Dong3}, VDSR \cite{Kim6}, GIF \cite{He19}, FGS \cite{Li24}, RGIF \cite{Ham29}, TGV \cite{Ferstl25}, RGDR \cite{Liu22}, HQDU \cite{Park21}, MRF \cite{Diebel20}, and our CDcGAN}
\label{fig::sharkdepth}
\vspace{-2mm}
\end{figure}

\begin{figure}[t]
\centering
\includegraphics[width=3in]{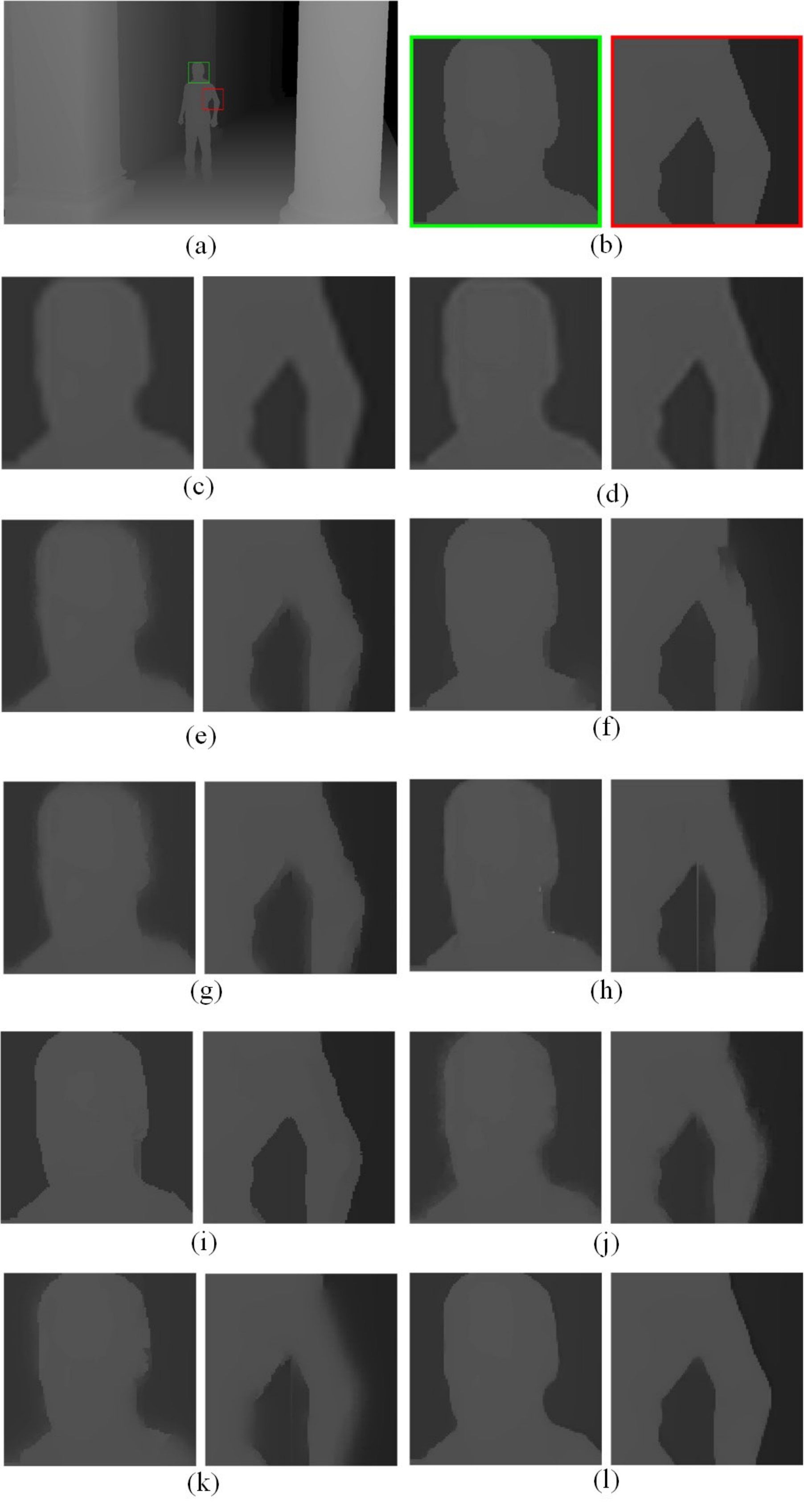}
\caption{\small The SR results for depth image with 4x scaling factor. (a) the first frame of Undo\_Dancer, (b) the close-ups of (a), (c-m) the close-ups of the results respectively with Bicubic interpolation, SRCNN \cite{Dong3}, VDSR \cite{Kim6}, GIF \cite{He19}, FGS \cite{Li24}, RGIF \cite{Ham29}, TGV \cite{Ferstl25}, RGDR \cite{Liu22}, HQDU \cite{Park21}, MRF \cite{Diebel20}, and our CDcGAN}
\label{fig::undodepth}
\vspace{-2mm}
\end{figure}

We compare the proposed approach with ten methods including Bicubic interpolation, SRCNN \cite{Dong3}, VDSR \cite{Kim6}, GIF \cite{He19}, FGS \cite{Li24}, RGIF \cite{Ham29}, TGV \cite{Ferstl25}, RGDR \cite{Liu22}, HQDU \cite{Park21}, MRF \cite{Diebel20}. The methods of SRCNN \cite{Dong3}, VDSR \cite{Kim6} only take the LR-depth image as input. For joint SR methods: GIF \cite{He19}, FGS \cite{Li24}, RGIF \cite{Ham29}, TGV \cite{Ferstl25}, RGDR \cite{Liu22}, HQDU \cite{Park21}, MRF \cite{Diebel20}, we use both low-resolution depth image and the ground-truth HR color image to get the results of depth super-resolution with the codes provided by the authors. As described above, our CDcGAN uses the low-resolution depth image and low-resolution color image as the input of our network. The objective quality comparison results for depth super-resolution are presented in the Table~\ref{tbl::depth-objectivemeasureX2}, ~\ref{tbl::depth-objectivemeasureX4}. From these tables, it can be found the PSNR, SSIM, and sharpness measurements of the proposed approach are better than SRCNN \cite{Dong3}. In addition, from the (d) of Fig. ~\ref{fig::bookdepth}, ~\ref{fig::sharkdepth}, ~\ref{fig::undodepth}, it can be clearly seen that there are severe artifacts existed in the SR image with SRCNN \cite{Dong3}, but proposed method does not suffer this problem. Although the performance on SSIM and PSNR of VDSR \cite{Kim6} is slight better than proposed method, our method has more sharpness than VDSR and the SR depth image of our CDcGAN looks more similar to the truth depth image, as displayed in the (e) of Fig. ~\ref{fig::bookdepth}, ~\ref{fig::sharkdepth}, ~\ref{fig::undodepth}. The depth sharpness profits depth image's applications, such as depth-based image rendering, scene's foreground extraction. The objective and subjective quality of the proposed method has better performance than several novel joint methods such as  optimization and filtering including GIF \cite{He19}, FGS \cite{Li24}, RGIF \cite{Ham29}, TGV \cite{Ferstl25}, RGDR \cite{Liu22}, HQDU \cite{Park21}, MRF \cite{Diebel20}, although these methods use HR color image. From the (f-m) of Fig. ~\ref{fig::bookdepth}, ~\ref{fig::sharkdepth}, ~\ref{fig::undodepth}, it can be found that most of these methods still has the problem of texture-copy and bleeding artifacts, due to depth SR problem's sensitivity to textural details and the weak boundary of color image.
\subsection{The visual comparison of architecture's application on image smoothing and edge detection}
\begin{figure}[t]
\centering
\includegraphics[width=3in]{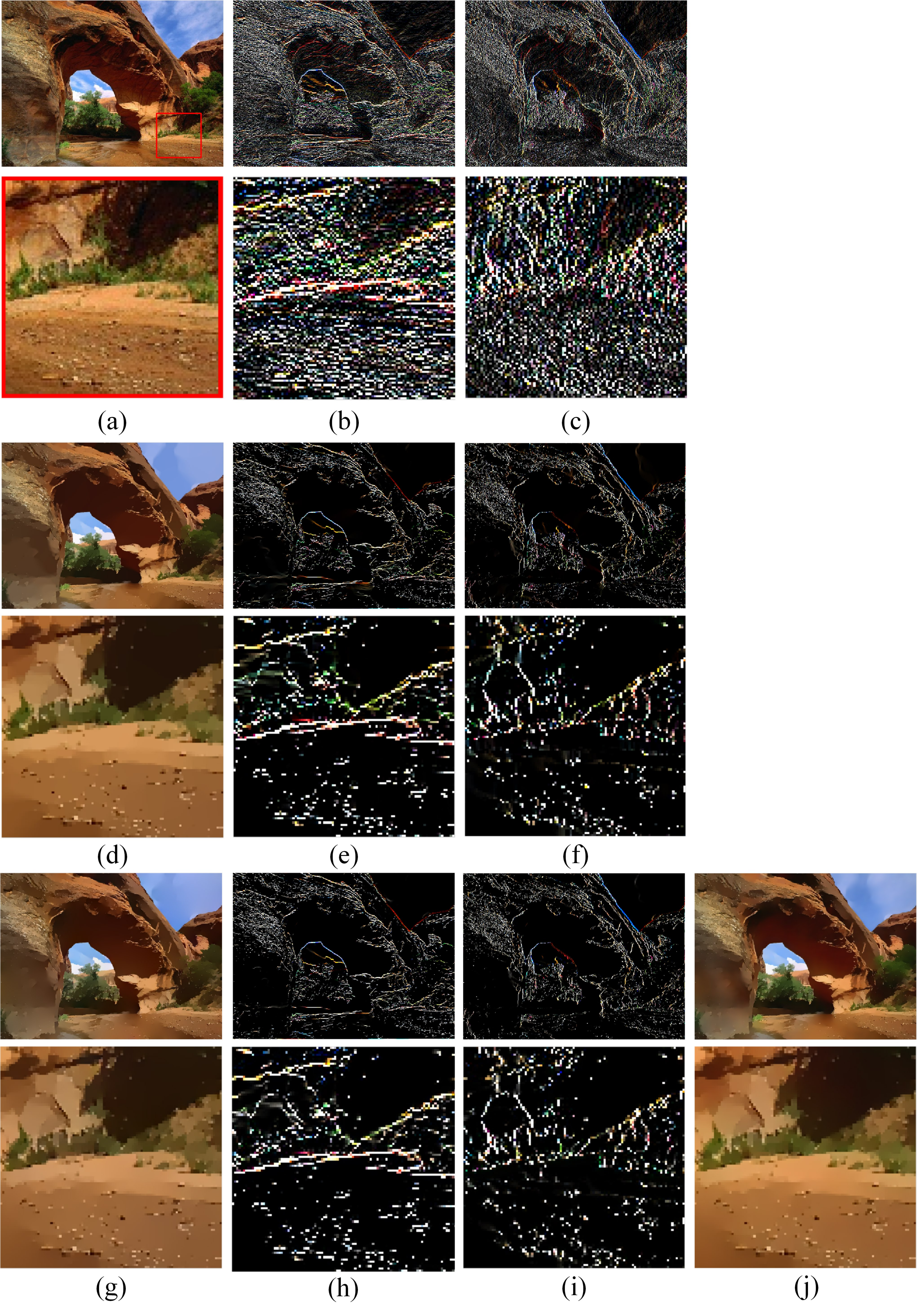}
\caption{\small The results of image smoothing and edge detection. (a-i) the first rows are the full-resolution image; the second rows are the close-ups located in the red rectangle; (a) the input image, (b-c) input image's edges of (a) in the horizonal and vertical direction, (d) the smoothed image with L0 gradient minimization approach \cite{Xus}, (e-f) the edges of (d) in the horizonal and vertical direction, (g) the smoothed image using output edges of the proposed network, (h-i) the output edges with the proposed network, (j) the smoothed image with the proposed network}
\label{fig::smoothing1}
\vspace{-2mm}
\end{figure}

\begin{figure}[t]
\centering
\includegraphics[width=3in]{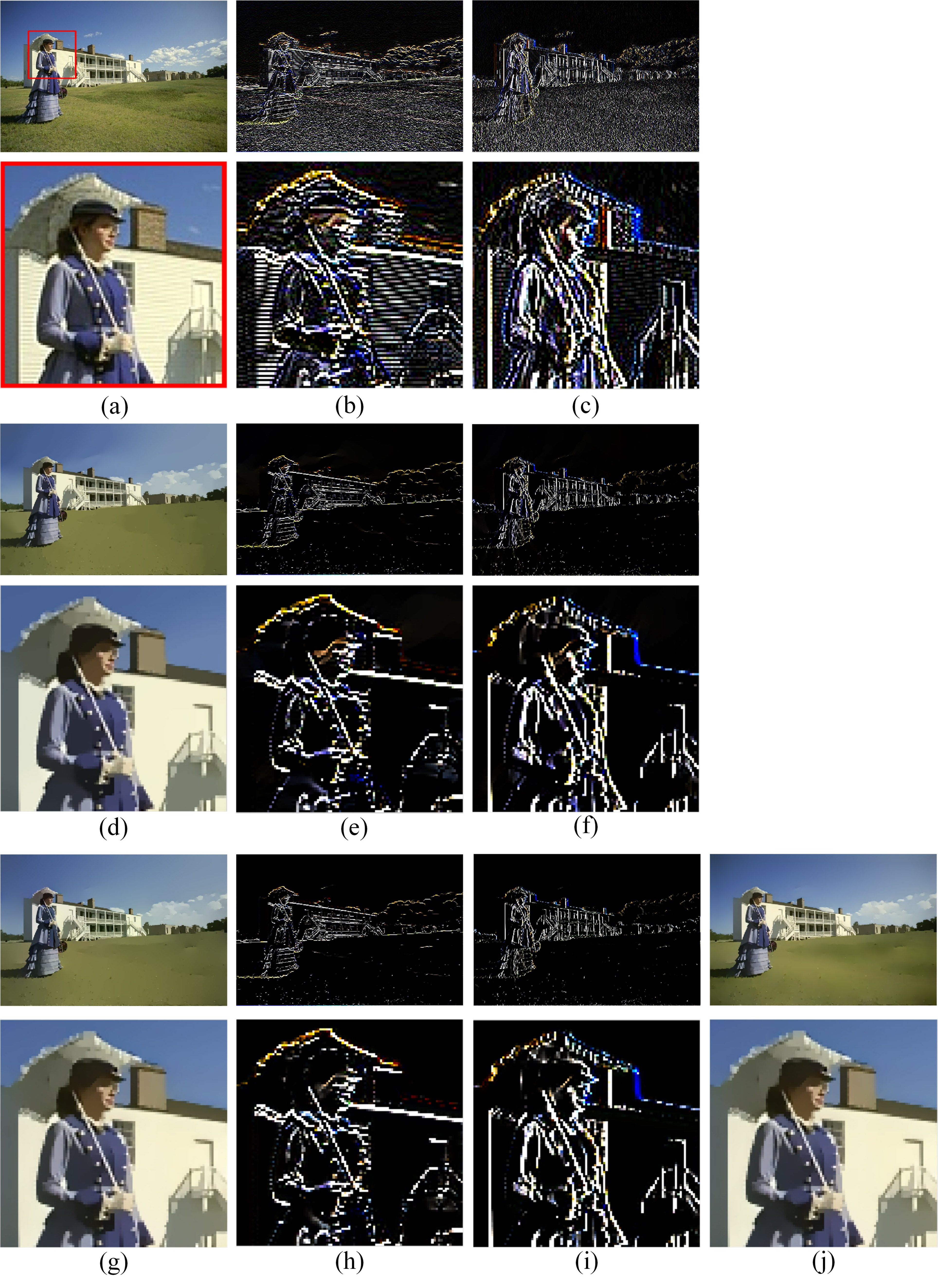}
\caption{\small The results of image smoothing and edge detection. (a-i) the first rows are the full-resolution image; the second rows are the close-ups located in the red rectangle; (a) the input image, (b-c) input image's edges of (a) in the horizonal and vertical direction, (d) the smoothed image with L0 gradient minimization approach \cite{Xus}, (e-f) the edges of (d) in the horizonal and vertical direction, (g) the smoothed image using output edges of the proposed network, (h-i) the output edges with the proposed network, (j) the smoothed image with the proposed network}
\label{fig::smoothing2}
\vspace{-2mm}
\end{figure}

In \cite{XL}, deep edge-aware filter is proposed to achieve the tasks of learning different image smoothing approaches such as L0 gradient minimization \cite{Xus}. This paper uses a deep convolutional neural network to learn various filtering in the gradient domain. Different from this paper, we use the proposed network to learn image smoothing filtering in both image domain and gradient domain to finish the tasks of image smoothing and edge detection. We use the learned gradient information for image smoothing in the gradient domain according to \cite{XL}, in which you can find the detail operations. As displayed in Fig~\ref{fig::smoothing1}(g-j) and Fig~\ref{fig::smoothing2}(g-j), we can see that our image smoothing results in both gradient domain and image domain are very close to the ones of L0 gradient minimization in the gradient domain \cite{Xus}. In \cite{Xus}, it has reported that there is some problems for their deep edge aware filters, such as unsatisfactory approximation to a few edge-preserving operators, during learning the filters in the image domain, but our architecture does not have this problem, due to the usage of TV loss, and gradient difference loss with L1 norm.  So the extensive application on the tasks of simultaneously image smoothing and edge detection have validated our architecture's flexibility and generality.

\section{Conclusion}
In this paper, color-depth conditioned generative adversarial network is trained to achieve color-depth super-resolution concurrently.  Three auxiliary losses are used as complementary regularization terms to train our networks in order to ensure the generated image close to the ground truth images, in addition to the adversarial loss. More importantly, we also apply our architecture to concurrently resolving the problems of image smoothing and edge detection.




\bibliographystyle{IEEEtran}
\bibliography{IEEEfull,cdganbibfile}
\end{document}